# PREDICTION OF MUSCLE ACTIVATIONS FOR REACHING MOVEMENTS USING DEEP NEURAL NETWORKS

Najeeb Khan and Ian Stavness

University of Saskatchewan, Saskatoon, SK, Canada
email: ian.stavness@usask.ca, web: http://biglab.ca/

## INTRODUCTION

The motor control problem involves determining the time-varying muscle activation trajectories required to accomplish a given movement. Muscle redundancy makes motor control a challenging task: there are many possible activation trajectories that accomplish the same movement. Despite this redundancy, most movements are accomplished in highly stereotypical ways. For example, point-to-point reaching movements are almost universally performed with very similar smooth trajectories [1].

Optimization methods are commonly used to predict muscle forces for measured movements [2]. However, these approaches require computationally expensive simulations and are sensitive to the chosen optimality criteria and regularization. Linear dimensionality reduction has also been proposed to identify low-dimensional motor modules that can account for stereotyped movements [3]. However, musculoskeletal systems are highly non-linear, making linear methods less reliable.

Deep neural networks (DNNs) are biologically inspired models that can be employed for non-linear dimensionality reduction. Deep autoencoders are DNN models that can automatically learn successively low dimensional features by transforming the input data through different layers of non-linearity. DNNs have been used to predict torque trajectories from initial and final state information in [4]. However, DNNs have yet to be applied to predict time varying muscle activations, in which the redundancy problem exists.

In this work, we investigate deep autoencoders for the prediction of muscle activation trajectories for point-to-point reaching movements. We evaluate our DNN predictions with simulated reaches and two methods to generate the muscle activations:

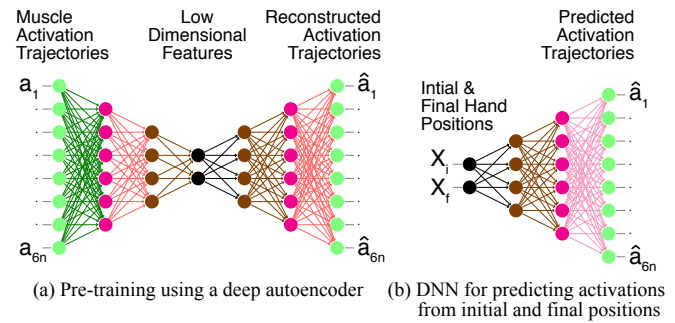

**Figure 1**: A deep autoencoder is trained to learn low-dimensional features that are used to map initial and final hand positions to muscle activations.

inverse dynamics (ID) and optimal control (OC) criteria. We also investigate optimal network parameters and training criteria to improve the accuracy of the predictions.

## METHODS

*Simulated reaches*: Random initial points for the end-effector of a two-link, six-muscle arm [5] were uniformly sampled in a 50 cm x 20 cm rectangular region in the hand-space. A random direction was chosen for each initial point and a final point for the reaching movement was uniformly selected in that direction within 10 cm. The data set consisted of 4500 pairs of initial/final points for training and 500 pairs for testing.

*Inverse Dynamics*: The initial and final points were connected by a time sampled minimum-jerk trajectory with a duration of 1 second. A sampling rate of 50 samples per second was used for sampling the min-jerk path. For each of the points in the minimum-jerk trajectory a torque control vector was calculated by using inverse kinematics and inverse dynamics of the arm model. 300-dimensional muscle activations for the six muscles were computed by minimizing the quadratic norm of the activations under the target torque constraint.

*Optimal Control*: For each pair of initial/final points, 50-dimensional torque control signals were generated for each joint using the iterative Linear-Quadratic-Gaussian (iLQG) method [5]. Muscle activations were computed using static optimization with a quadratic cost.

*Network Training*: For each control type, we trained an autoencoder with layer dimensions 300-150-50-4-50-150-300 (Figure 1a). Layer-wise pre-training [6] was used to train the network with the muscle activation trajectories as the inputs and outputs. The decoder part of the autoencoder (Figure 1b) with dimensions 4-50-150-300 was then retrained on the inputs as the initial/final end-effector positions, and outputs as the muscle activations. Both the networks were trained by minimizing the cross-entropy loss between predicted and desired output using the conjugate gradient descent method.

**RESULTS AND DISCUSSION**

Muscle activations predicted with the DNN matched well with the actual trajectories for both the ID and OC conditions (Figure 2). RMS error in the activation trajectories was 0.0048 and 0.0067 for the ID and OC reaches, respectively.

Simulated hand trajectories, when the arm model is driven with DNN-predicted muscle activations, also matched well with the original hand trajectories (Figure 3). The average error in reaching the target was 0.125 cm and 0.127 cm for the ID and OC reaches, respectively.

Previously, DNNs for predicting joint torques used mean squared error (MSE) as the training criteria and canonical optimization for minimization [4]. We found that using cross entropy loss instead of MSE as the training criteria and using conjugate gradient descent for minimization with mini-batch training outperformed [4]. Even with the problem extended from torques to activations, our predictions resulted in an average endpoint error of 0.127 cm, whereas [4] reported an endpoint error of 0.347 cm. Also, we performed a comprehensive evaluation of our methods and reported our results with 500 samples as compared to only 14 in [4].

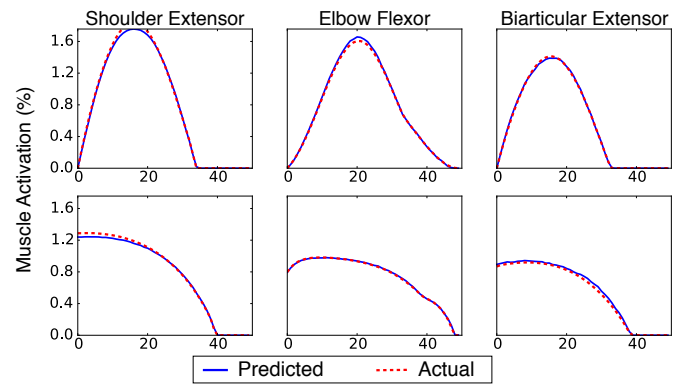

**Figure 2**: DNN predicted activations for 3 muscles for a reaching movement generated by inverse-dynamics (top) and optimal control (bottom).

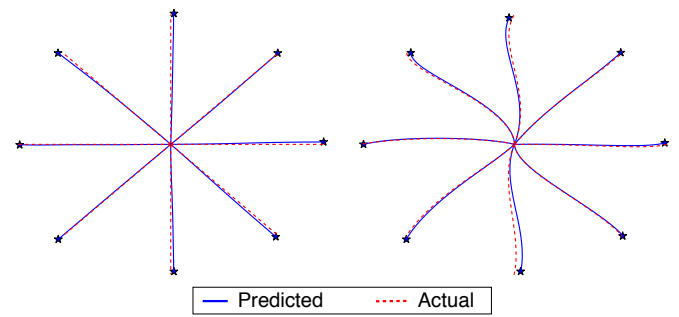

**Figure 3**: Hand trajectories (from center outward) driven by DNN predicted activations for inverse-dynamics (left) and optimal control (right).

**SUMMARY**

In this study, we demonstrated that with proper learning criteria and training methods, DNNs can accurately generate simulated muscle-driven reaches with both inverse-dynamics and optimal control based activation trajectories. Our work provides a proof of concept that DNNs can be used for more complex biomechanical models. As future work, we plan to investigate recurrent neural networks to model variable duration reaches.